\documentclass[runningheads]{llncs}
\usepackage[T1]{fontenc}
\usepackage{graphicx}
\usepackage[ruled]{algorithm2e}
\usepackage{helvet}  
\usepackage{courier}  
\usepackage{url}
\usepackage{amssymb}
\usepackage{amsmath}
\usepackage{bm}
\usepackage{booktabs}
\usepackage[capitalise]{cleveref}
\usepackage{color}
\usepackage{footnote}

\usepackage{subfigure}
\usepackage{parskip}
\usepackage{multirow}
\usepackage[title]{appendix}

\usepackage[normalem]{ulem}
\useunder{\uline}{\ul}{}
\usepackage{CJKutf8}

\begin{document}
%
%\title{Contribution Title\thanks{Supported by organization x.}}
\title{TegFormer: Topic-to-Essay Generation with Good Topic Coverage and High Text Coherence}
\titlerunning{TegFormer}
\author{Wang Qi\inst{1} \and
Rui Liu\inst{1} \and
Yuan Zuo\inst{2} \and
Yong Chen\inst{3}\thanks{Corresponding Author. Email:~alphawolf.chen@gmail.com.}
\and Dell Zhang\inst{4}}
%\author{Anonymous Author(s)}
%
% \authorrunning{Anonymous Submission}
\institute{School of Computer Science, Beihang University, Beijing, 100191, China \and
	School of Economics and Management, Beihang University, Beijing, 100191, China \and
School of Computer Science (National Pilot Software Engineering School), Beijing University of Posts and Telecommunications, Beijing, 100876, China
%\email{alphawolf.chen@gmail.com}\\
%\url{http://www.springer.com/gp/computer-science/lncs} 
\and
Thomson Reuters Labs, London, UK\\
}
\maketitle              % typeset the header of the contribution
\begin{abstract}
Creating an essay based on a few given topics is a challenging NLP task.
Although several effective methods for this problem, topic-to-essay generation, have appeared recently, there is still much room for improvement, especially in terms of the coverage of the given topics and the coherence of the generated text. 
In this paper, we propose a novel approach called TegFormer which utilizes the Transformer architecture where the encoder is enriched with domain-specific contexts while the decoder is enhanced by a large-scale pre-trained language model. 
Specifically, a \emph{Topic-Extension} layer capturing the interaction between the given topics and their domain-specific contexts is plugged into the encoder.
Since the given topics are usually concise and sparse, such an additional layer can bring more topic-related semantics in to facilitate the subsequent natural language generation.
Moreover, an \emph{Embedding-Fusion} module that combines the domain-specific word embeddings learnt from the given corpus and the general-purpose word embeddings provided by a GPT-2 model pre-trained on massive text data is integrated into the decoder.
Since GPT-2 is at a much larger scale, it contains a lot more implicit linguistic knowledge which would help the decoder to produce more grammatical and readable text. 
Extensive experiments have shown that the pieces of text generated by TegFormer have better topic coverage and higher text coherence than those from SOTA topic-to-essay techniques, according to automatic and human evaluations. 
As revealed by ablation studies, both the Topic-Extension layer and the Embedding-Fusion module contribute substantially to TegFormer's performance advantage.

\end{abstract}

\section{Introduction} 
\label{sect:intro}

\begin{figure}[t!]
	\centering
	\includegraphics[width=0.5\textwidth]{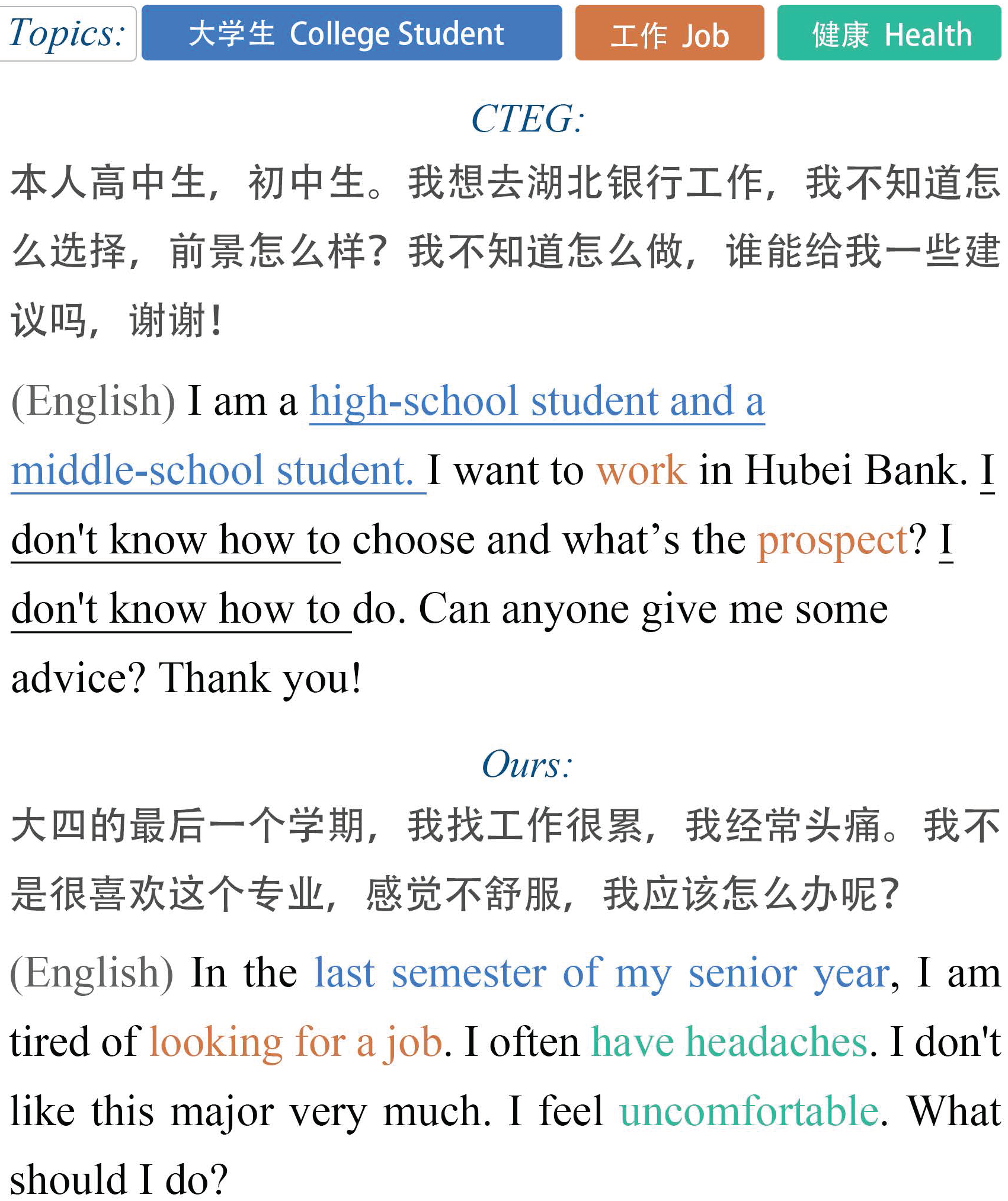}
	\caption{ 
		Two real essays generated by a strong model CTEG and our TegFormer method.
		In light of the final texts, we made a few marks on the English translations. 
		First, we use three colors to distinguish each topic, and dye the segment of text to the same color as the topic, of which the semantics are covered by that segment.
		Second, we underline the segment of text where there exist a problem of logic or repeat.
		Clearly, texts generated by CTEG not only suffer from logical contradiction, but also fail to cover the ``Health'' topic.
	}
	\label{fig:example}
\end{figure}

The task of \textbf{t}opic-to-\textbf{e}ssay \textbf{g}eneration~(TEG)~\cite{MTA-LSTM} is to let machines generate a coherent essay that covers a few given topics (in the form of keywords/keyphrases), as illustrated in Fig~\ref{fig:example}. 
Due to many potential real-life applications such as automated report generation, intelligent tutoring, and robot journalism~\cite{INLG}, TEG has attracted more and more attention from both academia and industry in recent years. 
As a form of creative writing~\cite{C-Text-Generation}, it is challenging for machines to produce a human-readable essay with good topic coverage and high text coherence because the number of input topics are often too small to provide sufficient semantic contexts that could be leveraged by computational models.

Recently, several papers addressing the TEG task have been published and made notable progress in automatic text generations.
In the seminal work from \cite{MTA-LSTM}, a novel MTA method~\cite{MTA-LSTM} is developed to utilize a coverage mechanism to balance the expressions of topics, and thus generate readable texts. 
%however, MTA only considers the interactive semantics of given topics and ignores topic contexts or linguistic grammars, which usually leaves the generated texts either topic inconsistent or logically fallacious. 
%Subsequent research focuses on introducing external knowledge base or internal topic graph to enhance topic coverage or text coherence.
\cite{CTEG} introduced commonsense knowledge and adversarial training to their CTEG model~\cite{CTEG}, and enhanced the generated texts' topic coverage.
%\citeauthor{SCKTG} integrate CVAE~\cite{CVAE-1, CVAE-2} of controlling sentiment and adversarial training of controlling topic coverage together~\cite{SCKTG}.
\cite{SCKTG} integrated sentiment, knowledge, and adversarial training with CVAE~\cite{CVAE-1,CVAE-2} as SCTKG~\cite{SCKTG}, and realized good topic coverage.
\cite{CKE} used various topic-related semantics via knowledge distillation~\cite{know-distill} and knowledge graph in their TEGKE approach~\cite{CKE}, and yielded high-quality and diverse essays.
To sum up, these research have tried their best to make full use of input topics either with extra knowledge (distillation/commonsense) or learning manners (adversarial training/Conditional VAE), and produced readable, topic coverage, or diverse texts.

However, as Fig.~\ref{fig:example} exhibits that given three topics, CTEG, a most competitive method, generated an essay with a logical fallacy (``\textbf{a high-school and middle school student}''), a repeated expression (``\textbf{i don't know how to do}''), and the ``\textbf{Health}'' topic not covered, which tells us that linguistic grammars are also essential for natural language generations (e.g., TEG) apart from the widely concerned semantics. Besides, methods like CTEG~\cite{CTEG} and SCTKG~\cite{SCKTG} adopt adversarial training, where the discriminator is a classifier over topics, limiting the number of topics to only 100 in Zhihu-Refined dataset\footnote{\url{https://pan.baidu.com/s/17pcfWUuQTbcbniT0tBdwFQ}\label{fn_zhihuRefined}}~\cite{CTEG}, far smaller than 5,559, the number of topics appeared in the original Zhihu dataset\footnote{\url{https://pan.baidu.com/s/1eC4gb_We33kr-ZbHn3KdIA}\label{fn_Zhihu}}~\cite{MTA-LSTM}. 
This implies that adversarial training would consume huge computing resources for robust results, and probably even fail on large-scale dataset.

To ensure good topic coverage and high text coherence in TEG, we propose a novel method called \textbf{TegFormer} which extends the Transformer architecture by enriching the encoder with more domain-specific contexts and enhancing the decoder with more background linguistic knowledge. 
The main contributions of this paper are as follows. 
\begin{itemize}
	\item We have designed the Topic-Extension layer which is plugged in the encoder seamlessly to capture the semantic interactions between the given topics and their domain-specific contexts. 
	It can bring in more topic-related semantics for the subsequent decoder, when the given topics are usually sparse in semantics.
	\item We have devised the Embedding-Fusion module which is plugged in the decoder seamlessly to capture the attentive sum of the domain-specific token embedding and the pre-trained GPT-2's embedding. 
	It can empower the decoder to generate more human-friendly texts since GPT-2 learns from large-scale open data and embeddings inferred from it implicitly contain abundant semantics and grammars.
	\item Extensive experimental comparisons between TegFormer and several state-of-the-art methods (including the pre-trained big models such as BART~\cite{BART} and T5~\cite{T5}) show that TegFormer performs best according to both automatic and human evaluation.
	Additional ablation studies confirm the effectiveness of Topic-Extension and Embedding-Fusion in boosting the quality of the generated text.
\end{itemize}

\section{Related Work}

\paragraph{Topic-to-Essay Generation.} Automatically generating an essay with several topics (i.e., TEG) is quite a difficult task in NLP. 
\cite{MTA-LSTM} use coverage vector to integrate topic information. 
\cite{CTEG} employ general knowledge to enrich the input semantics and adopt adversarial training to enhance topic coverage. \cite{SCKTG} inject the sentiment information into the generator for controlling sentiment based on CVAE and adversarial training. \cite{CKE} propose comprehensive knowledge enhanced model via knowledge distillation and knowledge graph.
These methods can generate either readable, topic coverage or diverse texts.
Besides, there are other relevant generation tasks such as controllable text generations, which demand on style/field/sentiment~\cite{PP-VAE,PPLM} control, and poetry generations~\cite{Poetry-1,Poetry-2} which require neural networks to compose poems with specific logics and rhymes given the input title. 
In this paper, we mainly focus on the topic-to-essay generation, whose success is up to not only semantic enhancements but also linguistic grammars.

\paragraph{Pre-trained Language Models.} 
There exist two types of pretrained language models (PLMs)~\cite{ZY}: discriminative PLMs such as BERT~\cite{BERT} and generative PLMs such as GPT~\cite{GPT-2,GPT-3}, BART~\cite{BART} and T5~\cite{T5}. 
The key difference lies in that there is a decoder in generative PLMs so that they can generate text sequences. 
The way of transferring a pre-trained generative PLMs to downstream tasks is to construct suitable inputs and outputs, and then fine-tune the models on the task-oriented datasets. 
Such a two-stage paradigm, i.e., pre-training plus fine-tuning, shows notable effects on most generation tasks. 
However, there also exist some issues, such as catastrophic forgetting, the task gap between the pretraining and finetuning stage, and no pertinent designs for downstream tasks~\cite{CF-1,CF-2}. 
Different from the above learning mode, TegFormer freezes the introduced GPT-2 and devises a Topic-Extension layer and an Embedding-Fusion module based on the characteristics of the TEG task.

\section{Problem Statement}
Currently, there are two representative understandings (i.e., automatic free-form texts and sequential storytelling) w.r.t. topic-to-essay generations, e.g., MTA~\cite{MTA-LSTM}, and Plan\&Write~\cite{Plan_And_Write}. 
Specifically, approaches like MTA view the input topics as a word set and let the machine produce a free-form text;
while methods like Plan\&Write treat them as a word sequence and let the machine generate a story.
In this paper, due to the properties of datasets and application scenarios, we formulate the task as a transition from a topic set to a paragraph-level text instead of the seq2seq task. 

Formally, given a dataset including pairs of a topic set $t^{1:N}=\{t_{1},\cdots,t_{N}\}$, and its related essay $y^{1:L}=\{y_{1},\cdots,y_{L}\}$, we want a $\bm{\theta}$-parameterized model to generate an essay which could well match its marked essay $y^{1:L}$ in terms of topic coverage and text coherence; in other words, there holds:
\begin{equation}
\hat{\bm{\theta}} = \arg \max_{\bm{\theta}}{\rm{P}}_{\bm{\theta}}(y^{1:L}|t^{1:N}), 
\end{equation}
where $\hat{\bm{\theta}}$ is the optimal parameter of the expected model $\rm{P}_{\bm{\theta}}(\cdot|\cdot)$. By the way, $L$ is the length of the generated text, and $N$ denotes the number of topic words, generally far smaller than $L$ (i.e., $N \ll L$).

\begin{figure*}[!tb]
	\center
	\includegraphics[width=0.98\textwidth]{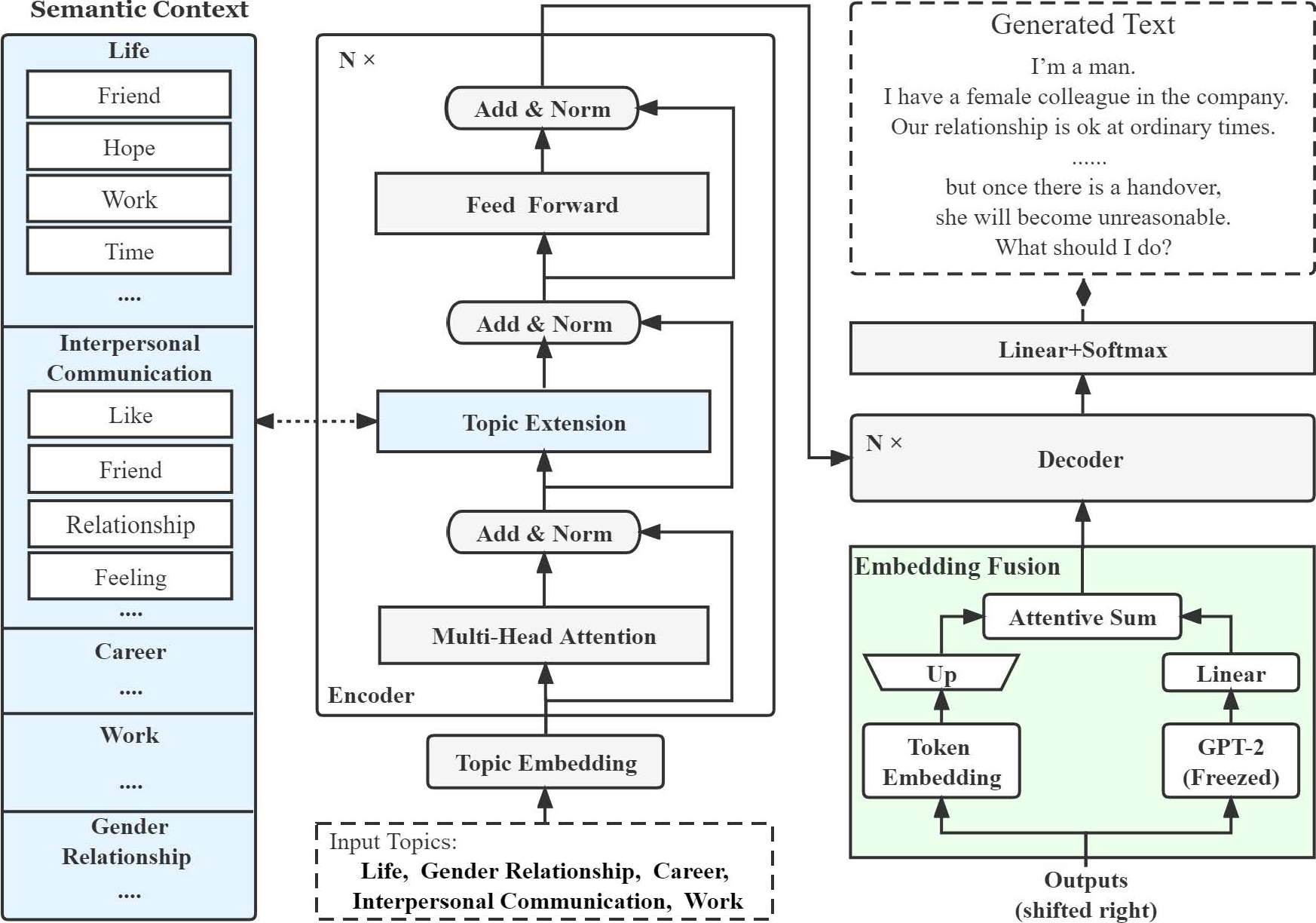}
	\caption{The \textbf{TegFormer} framework: the inputs are the given topics and each topic's top-$k$ most relevent keywords, and the output in each token-generated step is a probability over vocabulary. 
	}
	\label{fig:model}
\end{figure*}

\section{Methodology}

\subsection{The Overall Framework}
Different from previous models which usually use LSTM/GRU~\cite{LSTM,GRU} as encoders or decoders, we adopt Transformer \cite{Transformer} as our base architecture, generating sequential texts from given topic sets.
%We remove the position embedding of  Transformer's encoder while converting the input topics to topic embeddings. 
Considering the problem of sequences generated by Transformer models, which are poor in performance on topic coverage~\cite{PPLM,PLM-3,C-Text-Generation,T5}, we devise the Topic-Extension layer, i.e., a semantic interaction between given topics and its surrounded semantic contexts, which can be seamlessly plugged in the Transformer encoder (seen as \cref{fig:model}). 
To alleviate the logical problems (as illustrated in \cref{fig:example}), we marry both domain and general embeddings into the Transformer decoder (seen as \cref{fig:model}). 
Since our methed is based on Transformer with customized layer/module for solving the TEG task, we call it \textbf{TegFormer}.

\begin{algorithm}[!tb]
	\caption{Topic Extension}
	\label{alg:1}
	\LinesNumbered
	\KwIn{Corpus $\chi$, i.e., a set of (topics, essay) pairs; the pre-set number of topic neighbors $k$;}
	\KwOut{(topic, keywords) mappings $\psi$.}
	Initialize topic-keyword Co-occurrence matrix $O$=\textbf{0};
	
	\ForEach{$x \in \chi$}{
		topics, essay = $x$;
		
		keywords = Extractor(essay); {/* The TextRank~\cite{TextRank} algorithm provided by the Python package \texttt{jieba} is used to extract keywords from essays.*/}
		
		\ForEach{topic in topics}{
			\ForEach{keyword in keywords}{
				$O$[topic][keyword] += 1;  
			}
		}
		
	}
	
	%Filter low-frequency topics and keywords in $O$; 
	
	Keywords with the top-$k$ largest values for each topic in $O$ as $\psi$.		
\end{algorithm}

Concretely, given topics $t_{1},\cdots,t_{N}$ as the input of the Transformer encoder, we remove their position encodings and convert them into:
\begin{equation}
\textbf{M} = {\rm{Encoder}}(t^{1:N}),
\end{equation}
where $\textbf{M}$ represents the output hidden embeddings of the Transformer encoder from input topics. 
We then use $\textbf{M}$ and the generated tokens $y_{1},\cdots,y_{i-1}$ in former steps to unlock the output hidden states $\textbf{h}_i^{o}$ of the Transformer decoder in the current step $i$:
\begin{equation}
\textbf{h}_i^{o}  =  {\rm{Decoder}}(\textbf{M}, y^{1:(i-1)}).
\end{equation}

Finally, as a conventional operation, we leverage the combination of a linear and a softmax layer, to yield a probability distribution over the vocabulary, based on which the $i$-th token $y_{i}$ is sampled with:
\begin{equation}
{\rm{P}}(y_{i}|t^{1:N}, y^{1:(i-1)}) =  {\rm Softmax}(\textbf{W} \times {\textbf{h}}_{i}^{o} + \textbf{b}),
\end{equation}
where $\textbf{W}$ and $\textbf{b}$ are the TegFormer's learnable parameters.

\subsection{The Topic-Extension Encoder}
\label{sect:TGET}

The number of given topics ususally range from $2$ to $7$, which is naturally not semantically sufficient for text generations. 
To address this problem, we deliberately design a Topic-Extension layer, i.e., a semantic interaction between input topics and their semantic contexts, which is implanted between the self-attention (Multi-Head Attention) sublayer and the FFN (Feed Forward Network) sublayer of the Transformer encoder, illustrated in the left and middle parts of \cref{fig:model}. 
In light of the topics' contexts, we mainly search for their top-$k$ most frequently-co-occurrent keywords within all sentences of the given dataset (as specified in \cref{alg:1}).
Note that \cref{alg:1} is a pre-processing step: once the extracted keywords for each topic are obtained, they would be fixed and the corresponding vocabulary indexes would be supplied to the Topic-Extension layer. Therefore, it should not add much computational burden to training or inference.

Denote the output hidden states of multi-head attention sublayer about topic $i$ in Transformer encode-layer $l$ as $\textbf{h}_i^{(l)}$, and the extracted keywords (semantic contexts) for topic $i$ as ${w}_{t_i} = \{{w}_{t_i}^{1}, \cdots, {w}_{t_i}^{k}\}$. 
We then use $\textbf{e}_{j}$ to represent the input embedding of the topic $i$'s context-keyword $w^j_{t_i}$.
Based on the above, the Topic-Extension layer can be described as follows:
\begin{equation}
%\begin{aligned}
%&
{q}_i^{(l)} = \textbf{W}_q^{(l)} \times \textbf{h}_i^{(l)},%\\
%&
\textbf{k}_j^{(l)} = \textbf{W}_k^{(l)} \times \textbf{e}_{j},%\\
%&
\textbf{v}_j^{(l)} = \textbf{W}_v^{(l)} \times \textbf{e}_{j},
%\end{aligned}
\end{equation}
and
\begin{equation}
\alpha^{(l)}_{ij} = \frac{\exp(\textbf{q}_i^{(l)} \times \textbf{k}_j^{(l)} )}{ {\sum_{z \in {w}_{t_i}}  \exp( \textbf{q}_i^{(l)} \times \textbf{k}_z^{(l)}  ) } }, 
\end{equation}
where $\textbf{W}_q^{(l)}$, $\textbf{W}_k^{(l)}$, and $\textbf{W}_v^{(l)}$ are learnable parameters.

Next, we use an Add\&Norm layer to stimulate the final output embedding of the encoder layer $l$ about topic $i$, which is marked as $\widetilde{\textbf{h}}_i^{(l)}$:
\begin{equation}
\widetilde{\textbf{h}}_i^{(l)} = \text{LN}(\textbf{h}_i^{(l)}  + {\sum_{j \in w_{t_i}} (\alpha^{(l)}_{ij} \times \textbf{v}_{j})}).
\end{equation}

Since the semantically-coherent keywords for each topic are located within the given dataset, such keywords could be treated as domain-specific semantic contexts for the given topics. 
As is evidently shown in \cref{fig:model}, the Transformer encoder covers not only the self-attentions among the given topics, but also the cross-attentions between the given topics and their semantic contexts, which accomplishes the semantic enhancements of the few input topics for the subsequent Transformer decoder.

\subsection{The Embedding-Fusion Decoder}
\label{sec: EFET}
%Quality of texts generated by Transformer hurts from lack of overall logic and poor diversity of expression. We attribute this phenomenon basically to the small scale of label data itself, in which Transformer can not learn enough general knowledge, or usually called commonsense knowledge~\cite{Knowledge-1, Knowledge-2}. Collecting and labelling  more high-quality data are expensive and thus we ought to solve this problem in model aspect. 
%Notice that most existing methods for TEG task such as CTEG~\cite{CTEG}, Apex~\cite{Apex}, and TEGKE~\cite{CKE} enrich  
Note that the pre-trained language model~\cite{PLM-1,PLM-2} can assimilate both semantic and grammatical knowledge from  a large-scale corpus~\cite{PLM-1,PLM-3}, and thus we devise an Embedding-Fusion module, i.e., an attentive (or to say, a data-driven adaptive) sum of the previous already-generated domain-specific token embeddings and their corresponding GPT-2 embeddings, which is plugged in the Transformer decoder (seen as \cref{fig:model}).

Specifically, recall the already-generated sequential tokens $\{y_{1},\cdots,y_{i-1}\}$,
their general token embeddings are achieved by a frozen GPT-2\footnote{It is from \textbf{https://huggingface.co/uer/gpt2-chinese-cluecorpussmall} and is pre-trained on a 14GB dataset named CLUECorpusSmall~\cite{CLUECorpus2020}.} and a learnable linear layer which could transform the output dimension of GPT-2 into the input dimension of the original Transformer decoder (as illustrated in the right branch of \cref{fig:model}). Meanwhile, their domain embeddings are obtained by a Finder (i.e., a token-embedding lookup table) with an up projection to the input dimension of the original Transformer decoder (as illustrated in the left branch of \cref{fig:model}). %~\cite{Houlsby-Adapter, Adapter-Eff, AdapterHub}. 
The formulation is:
\begin{equation}
\label{LM}
\{\textbf{e}_g(y_{j})\}_{j=1}^{i-1} = \textbf{W} \times \text{LM}(y_{1},\cdots, y_{i-1}),
\end{equation}	
\begin{equation}\label{Finder}
\textbf{e}_d(y_{j}) = \textbf{W}_{\text{up}} \times \text{Finder}(y_{j}),
\end{equation}
\begin{equation}
\alpha_j = \text{sigmoid}(\textbf{v}^\top \times [\textbf{e}_g(y_{j}),  \textbf{e}_d(y_{j})]),
\end{equation}
and
\begin{equation}
\textbf{e}(y_{j}) =  \alpha_j~\textbf{e}_g(y_{j}) + (1 - \alpha_j)~\textbf{e}_d(y_{j}),
\end{equation}
where LM is the frozen pre-trained GPT-2 model and $\text{Finder}(y_{j})$ marks a lookup function that can locate a token's data-driven to-be-learnt embedding, \textbf{W} and $\textbf{W}_{\text{up}}$ are two trainable linear layers.

Since $\textbf{e}_d(y_j)$ is a to-be-learnt embedding based on the given text dataset, and $\textbf{e}_g(y_j)$ is a pre-trained model's embedding based on the open large-scale corpus, the Embedding-Fusion module enhances the Transformer decoder with domain and general semantics, and also with implicit linguistic grammars (GPT-2).

\subsection{Training and Inference}
We freeze the parameters of GPT-2 and adopt the Label Smoothing (LS in \cref{eq_8}) trick into a KL divergence loss, which is written as:
\begin{equation}
\label{eq_8}
\begin{split}
{\rm{D}}(\bm{\theta}) = \text{KL}({\rm{P}_{\bm{\theta}}}(y_{i}|y^{1:i-1}, t^{1:N})||\text{LS}(\tilde{y}_{i}) ),   
\end{split}
\end{equation}
where $\tilde{y}_{i}$ denotes the $i$-th word of the labelled essay.

As \cite{Repeat-analysis} analysed that the ``High Inflow'' problem is the major reason of the common repetition phenomenon in text generation tasks.
To solve this issue, we use top-$k$ sampling~\cite{Topk} for inference.

\section{Experiments}
%We validate the effectiveness of the proposed method on three real-world datasets. Seven state-of-the-art baseline methods are included for a thorough comparative study.

\begin{table*}[!tb]
	\small
	\centering
	%\footnotesize
	\setlength{\tabcolsep}{3.0mm}{
		\begin{tabular}{@{}l|rrrrrr@{}}
			\toprule
			\textbf{Dataset} & \textbf{Total} & \textbf{Train} & \textbf{Test} & \textbf{Topics} & \textbf{Avg-T} & \textbf{Avg-Len} \\ 
			\midrule
			\textbf{Essay} & 494,944 & 300,000 & 5,000 & 7,995 & 5.00 & 64.13 \\
			\textbf{Zhihu} & 56,221 & 50,000 & 5,000 & 5,559 & 5.00 & 78.14 \\
			\textbf{Zhihu-Refined} & 30,000 & 27,000 & 2,300 & 100 & 2.36 & 71.94 \\ 
			\bottomrule
	\end{tabular}}
	\caption{The dataset statistics: \textbf{``Total/Train/Test''} represents the number of essays in the whole/training/testing dataset. 
		\textbf{``Topics''} stands for the number of the topics appeared in the whole dataset. 
		\textbf{``Avg-T''} denotes the average number of topics for each essay. 
		\textbf{``Avg-Len''} refers to the average number of words within each essay.}
	\label{tab:data_stat}
\end{table*}

\subsection{Datasets}
Three public datasets have been used in our experiments as depicted in \cref{tab:data_stat}, where \textbf{Essay}\footnote{\url{https://pan.baidu.com/s/1_JPh5-g2rry2QmbjQ3pZ6w}} and \textbf{Zhihu}\textsuperscript{\ref{fn_Zhihu}} were both constructed by \cite{MTA-LSTM}, 
while \textbf{Zhihu-Refined}\textsuperscript{\ref {fn_zhihuRefined}} was constructed by \cite{CTEG} to replace the Zhihu corpus in their experiments. 
All these datasets are paragraph-level Chinese essays.

\begin{table*}[!tb]
	\small
	\centering
		\begin{tabular}{@{}l|l|cccc|cccc@{}}
			\toprule
			&  & \multicolumn{4}{c|}{Automatic Evaluation} & \multicolumn{4}{c}{Human Evaluation} \\ 
			\textbf{Dataset} & \textbf{Method} & \textbf{BLEU-2} & \textbf{Rouge-L} & \textbf{Dist-1} & \textbf{Dist-2} & \textbf{Cov.} & \textbf{Nov.} & \textbf{Log.} & \textbf{Coh.} \\ 
			\midrule
			\multirow{7}{*}{\textbf{Essay}} & MTA & 4.04 & 9.32 & 2.11 & 11.08 & 2.71 & 3.52 & 3.30 & 3.30 \\
			& Plan\&Write & 4.77 & 11.89 & 3.40 & 24.34 & 2.10 & 3.22 & 2.17 & 2.58 \\
			& Apex & 5.81 & 12.67 & 4.14 & 28.81 & 4.16 & 3.41 & 3.21 & 3.42 \\
			& Transformer & 5.78 & 11.35 & 4.02 & 28.42 & 3.80 & 3.51 & 2.62 & 3.18 \\
			& BART & 4.79 & 10.02 & 4.34 & 27.58 & 2.91 & 3.31 & 3.29 & 3.50 \\
			& T5 & 5.89 & 12.91 & 4.36 & 28.04 & 4.18 & 3.43 & 3.36 & 3.48 \\
			& \textbf{TegFormer} & \textbf{6.32} & \textbf{14.24} & \textbf{4.64} & \textbf{30.86} & \textbf{4.23} & \textbf{3.59} & \textbf{3.42} & \textbf{3.87} \\ \midrule
			\multirow{7}{*}{\textbf{Zhihu}} & MTA & 2.12 & 7.37 & 0.92 & 4.86 & 1.85 & 2.90 & 2.20 & 2.85 \\
			& Plan\&Write & 2.42 & 8.81 & 3.48 & 24.26 & 2.00 & 2.85 & 1.90 & 2.25 \\
			& Apex & 3.42 & 11.30 & 4.37 & 29.67 & 2.81 & 3.21 & 3.09 & 3.27 \\
			& Transformer & 2.91 & 10.60 & 4.00 & 29.39 & 2.39 & 2.85 & 2.05 & 2.25 \\
			& BART & 2.55 & 9.29 & 3.49 & 27.60 & 2.52 & 2.87 & 2.98 & 3.02 \\
			& T5 & 3.41 & 11.31 & 4.34 & 30.16 & 2.87 & 3.12 & 3.11 & 3.24 \\
			& \textbf{TegFormer} & \textbf{3.53} & \textbf{12.21} & \textbf{4.61} & \textbf{32.67} & \textbf{2.91} & \textbf{3.45} & \textbf{3.25} & \textbf{3.45} \\ 
			\midrule
			\multirow{2}{*}{\textbf{Zhihu-Refined}} & CTEG & 3.62 & 11.92 & 5.21 & 25.91 & 2.34 & 2.70 & 2.50 & 2.65 \\
			& \textbf{TegFormer} & \textbf{3.81} & \textbf{13.02} & \textbf{5.81} & \textbf{33.62} & \textbf{2.95} & \textbf{3.21} & \textbf{3.75} & \textbf{3.48} \\ 
			\bottomrule
	\end{tabular}%}
	
	\caption{The automatic and human evaluations for different methods on three real-world datasets. According to the t-test, the improvements brought by TegFormer over all baselines are significant at the level $p<0.05$.
	}
	\label{tab:main_res}
\end{table*}

\subsection{Baseline Methods}
Following are seven methods selected as baselines for comparable experiments.
\begin{itemize}
	\item \textbf{MTA}~\cite{MTA-LSTM} utilizes an LSTM decoder and a topic coverage vector to balance the expression of all topic information during generation.
	
	\item \textbf{Plan\&Write}~\cite{Plan_And_Write} applies storyline as the intermediate process of story generation. The framework uses BiLSTM as encoder and LSTM as decoder.
	
	\item \textbf{CTEG}~\cite{CTEG} introduces memory mechanism to store commonsense knowledge and leverages adversarial training to improve generation. It also uses LSTM as backbone of its framework.
	
	\item \textbf{Transformer}~\cite{Transformer} is implemented as a Seq2Seq baseline. 
	To be specific, we use a special token to separate sequential topic words as its input.
	
	\item \textbf{BART}~\cite{BART} is a Seq2Seq task oriented pre-trained model, which tries to re-construct the original texts from the corrupted ones. 
	We finetune it on our datasets.
	
	\item \textbf{T5}~\cite{T5} is also a Seq2Seq task oriented pre-trained model, which views all NLP problems as Seq2Seq tasks, and transfers tasks such as NMT~\cite{NMT}, text classification~\cite{classification}, and summarization~\cite{summarization} to Seq2Seq formats. 
	We finetune it on our datasets. 
	
	\item \textbf{Apex}~\cite{Apex} mixes the CVAEs model with a PI controller, where P-term is proportional to KL divergence loss and I-term is proportional to the integral of it, aiming to manipulate the diversity and accuracy of generated texts in E-commerce from the view of PID Control Algorithm~\cite{PID-1,PID-2}. 
	We follow~\cite{Apex} and set PI-$v$ to $2$.
	
\end{itemize}

Considering that the discriminator of CTEG is a multi-label classifier~(i.e., each topic is treated as a label), and thus CTEG has strict requirements on its training data, {in other words,} only the most frequent topics are involved. 
To make experiments go smoothly, CTEG builds the~\textbf{Zhihu-Refined} (much smaller than \textbf{Zhihu}) dataset. 
Following the conventions, we conduct experiments on \textbf{Zhihu-Refined} between TegFormer and CTEG, and make comparisons with other six baselines on \textbf{Essay} and \textbf{Zhihu}.

\subsection{Implementation Details}
We train our method for $120$ epochs and the batch size is 128. The optimizer is Adam~\cite{Adam} with learning rate set to $10^{-3}$. 
We use 6-layer encoders and decoders; the hidden layer size is 512 nodes with 8 attention heads; the intermediate size of FFN is 2048. Furthermore, to minimize the influence of model size and focus the comparison on model architecture, we use baselines of similar size: T5-small (6-layer), BART-base (6-layer) and Apex (with 6-layer Transformer). 
Our GPU is Tesla V100. When training Transformer~(base) or finetuning BART and T5, we split the input topics with a special token. 
We set 150 as the maximum length of the generated essays. 
The inference model stops when the end token~({ i.e.}, [EOS] or [SEP]) is encountered in the process of generation. 
More details can be found in the code to be released to the public. 

\subsection{Evaluation Metrics}
Both automatic and human evaluations are adopted to compare various methods' performance. 
For the automatic evaluation, we employ the popular metrics: Dist-1/2~\cite{CTEG}, ROUGE-L~\cite{Rouge-L}, and BLEU~\cite{BLEU}.

Following the popular settings of the human evaluation~\cite{CTEG,SCKTG,CKE},
we sample $200$ experimental results with each containing the given topics and their generated essays of different models. 
Then we invite $3$ annotators who don't know which model the generated essays come from, and ask them to score them from 1 to 5 in four criteria, i.e., coherence (\textbf{``Coh.''}), topic coverage (\textbf{``Cov.''}), novelty (\textbf{``Nov.''}) and logic (\textbf{``Log.''}). 
Finally, each method's score on a criterion is collected by averaging the scores of the three annotators. Notice that the Spearman's rank correlations between different human annotators' scores are high above $0.7$.

\subsection{Main Results}

\cref{tab:main_res} exhibits all the experimental results on three corpus. It's clear that no matter what kind of evaluations, \textbf{TegFormer} yields the highest scores, which proves our method's effectiveness on topic-to-essay generations.

By comparing several pre-trained language models (PLMs) such as BART\cite{BART}, Transformer~\cite{Transformer}, and T5~\cite{T5}, with the TEG task oriented models such as MTA~\cite{MTA-LSTM}, CTEG~\cite{CTEG} and Apex~\cite{Apex}, PLMs could catch up with the TEG approaches; 
this is mainly due to PLMs' rich semantics and grammars learned from large-scale datasets. 
However, the PLMs are still inferior to TegFormer, which implies that the gap between distinct tasks (pre-trained on one task, but learned for another task) limits their potentials to the fullest.

When it comes to TegFormer and Transformer, we could conclude that the Topic-Extension layer and the Embedding-Fusion module do promote the generated essays' topic coverage and text coherence.  

Besides, it's reported that CTEG defeats several previous methods on \textbf{Zhihu-Refined}~\cite{CTEG}, but this refined Zhihu dataset is quite small and not practical in real-life scenarios; 
and hence we just draw a comparison between our method and CTEG on \textbf{Zhihu-Refined}. 
As exhibited in \cref{tab:main_res}, our TegFormer method still outperforms CTEG on all evaluation metrics.

\subsection{Hyperparameter Analysis}
\begin{figure*}[htb]
	\centering
	\subfigure[The dimension of token embedding $\text{dim}(\text{Finder}(y_{j}))$.]{
		\label{fig:dim}
		\includegraphics[width=0.46\textwidth]{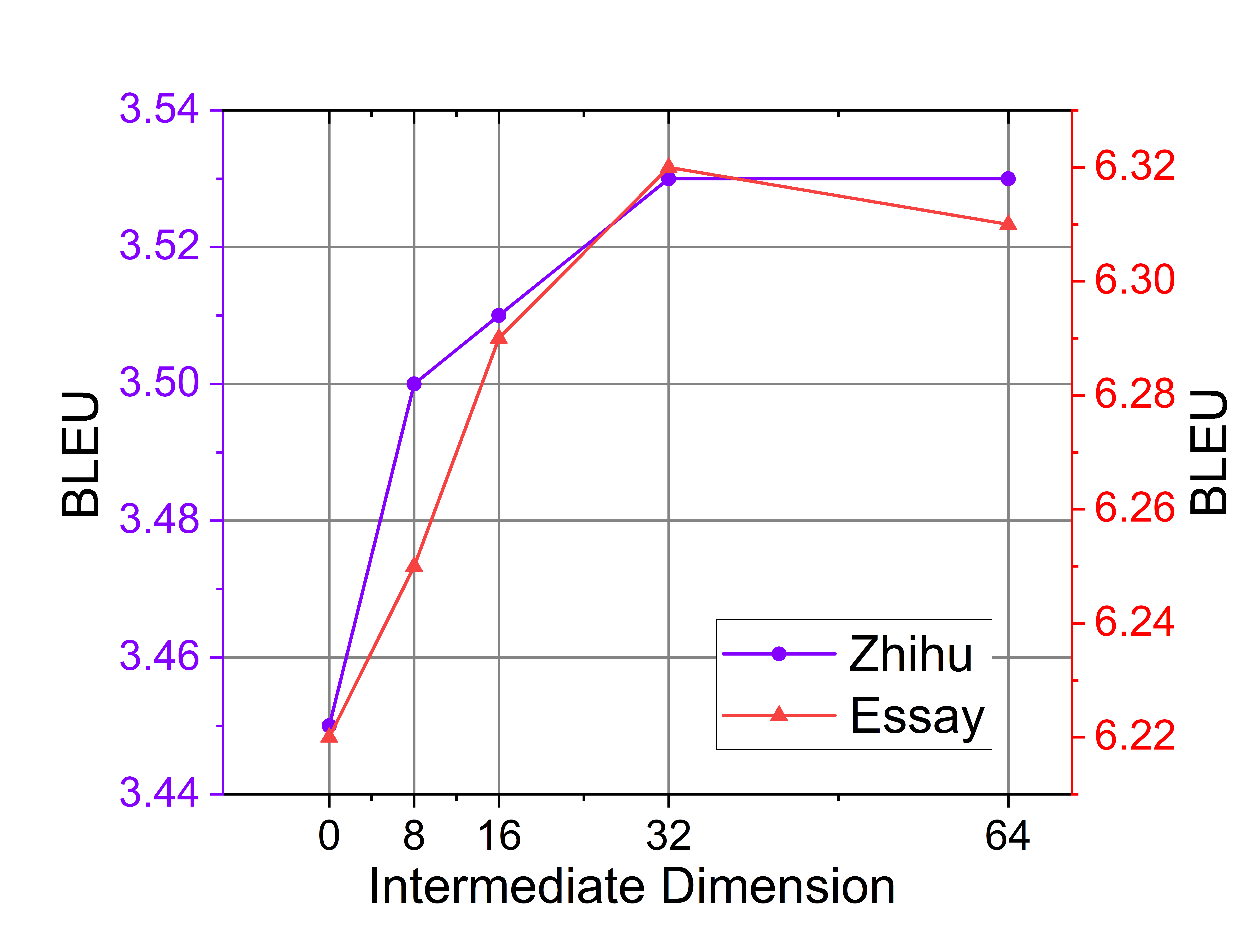}
	}
	\subfigure[The number of extended keywords $k$.]{
		\label{fig:k}
		\includegraphics[width=0.46\textwidth]{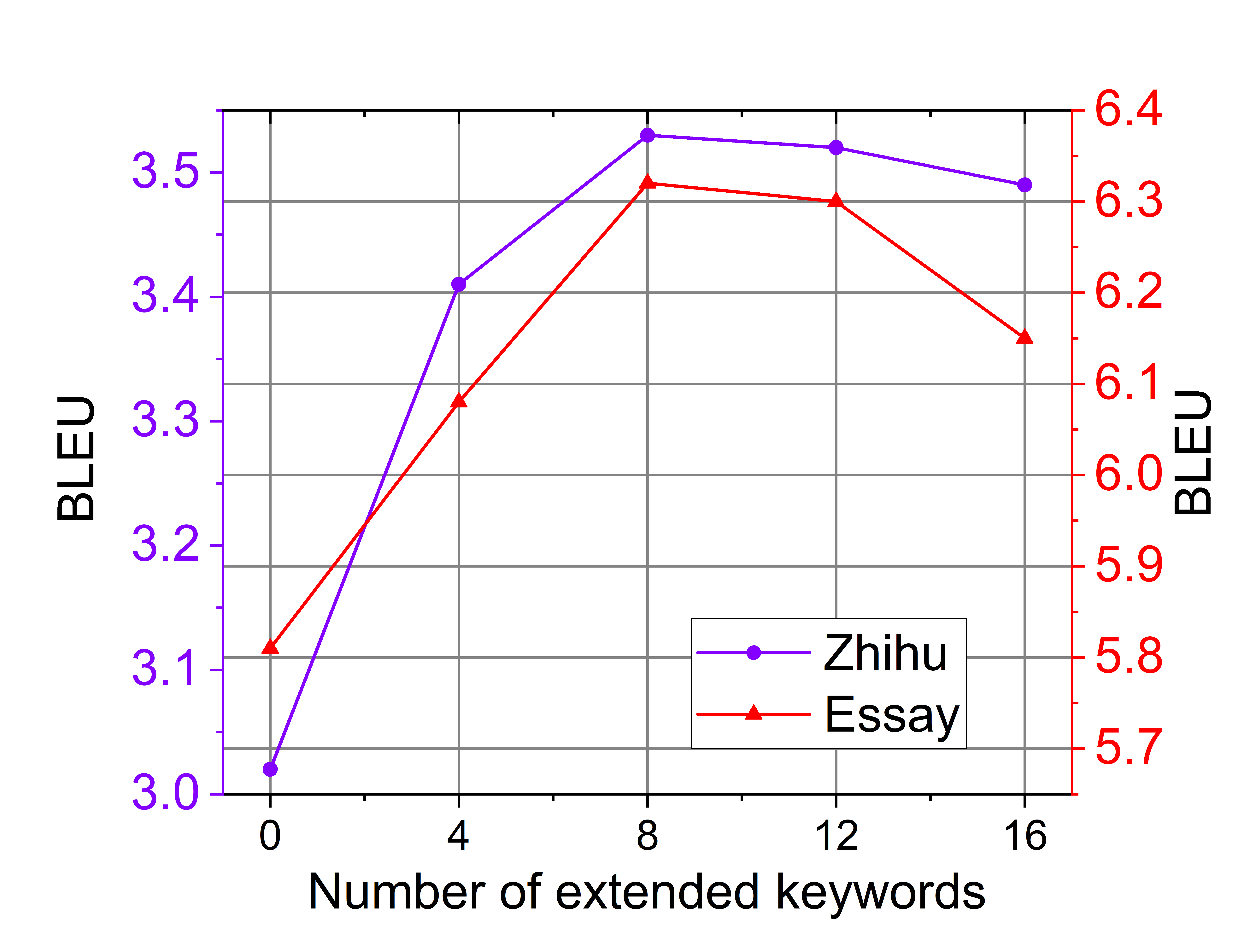}		 
	}
	\caption{BLEU scores of our TegFormer with different number of extended keywords in the Topic-Extension layer, and different dimensions of token embeddings in the Embedding-Fusion module, on \textbf{Zhihu} and \textbf{Essay} datasets.}
	\vspace{-0.3cm}
	\label{fig:agg_model}	
\end{figure*}

TegFormer owns two important hyper-parameters. 
The first is the dimension of the token embedding in the devised Embedding-Fusion module; 
as \cref{fig:dim} shows,
$\text{dim}(\text{Finder}(y_{j}))=32$ in \cref{Finder} leads to competitive results with that of larger dimensions. 
The second is the number of the extended keywords for each topic in the Topic-Extension layer; 
as can be seen from \cref{fig:k}, $k=8$ (i.e., taking the top-$8$ most relevant keywords for each topic) is suitable to extract the semantic contexts of their topics.    

\begin{table*}[!tb]
	\small
	\centering
	\setlength{\tabcolsep}{1.0mm}{
		\begin{tabular}{@{}l|l|cccc|cccc@{}}
			\toprule
			&  & \multicolumn{4}{c|}{Automatic Evaluation} & \multicolumn{4}{c}{Human Evaluation} \\ 
			\textbf{Dataset} & \textbf{Method} & \textbf{BLEU} & \textbf{Rouge-L} & \textbf{Dist-1} & \textbf{Dist-2} & \textbf{Cov.} & \textbf{Nov.} & \textbf{Log.} & \textbf{Coh.} \\
			\midrule
			\multirow{4}{*}{\textbf{Essay}} & Transformer & 5.78 & 11.35 & 4.02 & 28.42 & 3.80 & 3.51 & 2.62 & 3.18 \\
			& TegFormer$_{EF}$ & 6.22 & 13.77 & 4.43 & 29.04 & 4.21 & 3.50 & 2.91 & 3.41 \\
			& TegFormer$_{TE}$ & 5.81 & 11.91 & 4.53 & 30.42 & 4.08 & 3.53 & 3.36 & 3.78 \\
			& \textbf{TegFormer} & \textbf{6.32} & \textbf{14.24} & \textbf{4.62} & \textbf{30.79} & \textbf{4.23} & \textbf{3.59} & \textbf{3.42} & \textbf{3.87} \\ 
			\midrule
			\multirow{4}{*}{\textbf{Zhihu}} & Transformer & 2.91 & 10.60 & 4.00 & 29.39 & 2.39 & 2.85 & 2.05 & 2.25 \\
			& TegFormer$_{EF}$ & 3.45 & 12.14 & 4.36 & 29.81 & 2.88 & 2.98 & 2.21 & 2.41 \\
			& TegFormer$_{TE}$ & 3.02 & 10.72 & 4.57 & 32.43 & 2.57 & 3.36 & 2.89 & 3.26 \\
			& \textbf{TegFormer} & \textbf{3.53} & \textbf{12.21} & \textbf{4.60} & \textbf{32.48} & \textbf{2.91} & \textbf{3.45} & \textbf{3.25} & \textbf{3.45} \\ 
			\bottomrule
	\end{tabular}}
	\caption{The automatic and human evaluation results of ablation studies.}
	\label{tab:abl_study}
\end{table*}

\subsection{Ablation Study}
We have carried out ablation studies by comparing TegFormer with its incomplete variants including Transformer (Base), TegFormer w/o Embedding-Fusion (TegFormer$_{EF}$), and TegFormer w/o Topic-Extension (TegFormer$_{TE}$). 
As can be seen from \cref{tab:abl_study}, TegFormer achieves the best results according to all the evaluations. 
Obvious improvements on the logic and diversity from the vanilla Transformer to TegFormer (w/o Topic-Extension) demonstrate the effectiveness of the Embedding-Fusion module. 
Similarly, increased scores on topic coverage from the vanilla Transformer to TegFormer (w/o Embedding-Fusion) testify that the proposed Topic-Extension layer could further enhance the quality of generated texts. 

%\subsection{Case Study}

\section{Conclusion}

In this paper, we propose the TegFormer model --- a customized Transformer with Topic-Extension and Embedding-Fusion for TEG tasks. 
In the encoder, to alleviate the semantic sparsity of input topics, we have designed a built-in Topic-Extension layer which enhances the topic coverage by extending the semantic contexts of the given topics with automatically extracted keywords.  
In the decoder, we have developed a plug-and-play Embedding-Fusion module which blends domain-specific and general-purpose input embeddings to improve the text coherence of the generated essays. 
Comparing our proposed TegFormer with several generative pre-trained models and the SOTA approaches to TEG tasks, we have found that while either the Topic-Extension layer or the Embedding-Fusion module alone is able to increase the output quality, their ``marriage'' leads to the maximum performance boost for TEG.

%\subsubsection{Acknowledgements} Please place your acknowledgments at
%the end of the paper, preceded by an unnumbered run-in heading (i.e.
%3rd-level heading).

\newpage

\bibliographystyle{splncs04}
\bibliography{anthology}
 
\end{document}